\documentclass{article} 
\usepackage[final]{style/neurips_2020}
\usepackage{times}

\usepackage{amsmath,amsfonts,bm}

\def\eqref#1{equation~\ref{#1}}

\def\1{\bm{1}}

\DeclareMathAlphabet{\mathsfit}{\encodingdefault}{\sfdefault}{m}{sl}
\SetMathAlphabet{\mathsfit}{bold}{\encodingdefault}{\sfdefault}{bx}{n}
\newcommand{\tens}[1]{\bm{\mathsfit{#1}}}

\newcommand{\etens}[1]{\mathsfit{#1}}

\usepackage[utf8]{inputenc} 
\usepackage[T1]{fontenc}    
\usepackage{url}            
\usepackage{booktabs}       
\usepackage{amsfonts}       
\usepackage{nicefrac}       
\usepackage{microtype}      
\usepackage{xcolor}         
\usepackage{fancyhdr}
\DeclareMathAlphabet      {\mathbf}{OT1}{cmr}{bx}{n}
\usepackage{formatting/packages}
\usepackage{formatting/names}
\usepackage{formatting/macros}
\usepackage{formatting/results}
\usepackage[colorlinks=true,allcolors=blue]{hyperref}       

\usepackage[ruled,vlined,noend,linesnumbered]{algorithm2e}
\usepackage{xcolor}

\SetAlFnt{\footnotesize}

\SetCommentSty{xCommentSty}

\SetNlSty{mynlfont}{}{} 

\LinesNumbered

\SetSideCommentRight

\DontPrintSemicolon

\RestyleAlgo{algoruled}

\title{Apollo: Transferable Architecture Exploration}

\fancypagestyle{firststyle}
{
   \fancyhf{}
   \fancyfoot[C]{\footnotesize Page \thepage\ of \pageref{LastPage}}
}

\author{Amir Yazdanbakhsh\quad~Christof Angermueller\quad~Berkin Akin\quad~Yanqi Zhou
\And
Albin Jones\quad~Milad Hashemi\quad~Kevin Swersky\\
\And
Satrajit Chatterjee~\quad~Ravi Narayanaswami~\quad~James Laudon\\
\\Google Research\\
\texttt{\{ayazdan,christofa,bakin,yanqiz,alj\}@google.com}\\
\texttt{\{miladh,kswersky,schatter,swamiravi,jlaudon\}@google.com}
}

\usepackage[subtle]{savetrees}

\SetKwComment{Comment}{\textcolor{black}{$\triangleright$ }}{}

\usepackage{cleveref}
\crefformat{section}{\S#2#1#3}
\begin{document}
\newcolumntype{?}{!{\vrule width 0.5pt}}
\maketitle
\begin{abstract}
\label{sec:abstract}
The looming end of Moore's Law and ascending use of deep learning drives the design of custom accelerators that are optimized for specific neural architectures.
Architecture exploration for such accelerators forms a challenging constrained optimization problem over a complex, high-dimensional, and structured input space with a costly to evaluate objective function. 
Existing approaches for accelerator design are sample-inefficient and do not transfer knowledge between related optimizations tasks with different design constraints, such as area and/or latency budget, or neural architecture configurations.
In this work, we propose a transferable architecture exploration framework, dubbed \apollo, that leverages recent advances in black-box function optimization for sample-efficient accelerator design.
We use this framework to optimize accelerator configurations of a diverse set of neural architectures with alternative design constraints. We show that our framework finds high reward design configurations (up to 24.6$\%$ speedup) more sample-efficiently than a baseline black-box optimization approach.
We further show that by transferring knowledge between target architectures with different design constraints, \apollo is able to find optimal configurations faster and often with better objective value (up to 25$\%$ improvements).
This encouraging outcome portrays a promising path forward to facilitate generating higher quality accelerators.
\end{abstract}

\section{Introduction}
\label{sec:intro}
The ubiquity of customized accelerators demands efficient architecture exploration approaches, especially for the design of neural network accelerators.
However, optimizing the parameters of accelerators is daunting optimization task that generally requires expert knowledge~\cite{dfo:siam:2009,prac_dse:mascots:2019}.
This complexity in the optimization is because the search space is exponentially large while the objective function is a black-box and costly to evaluate. 
Constraints imposed on parameters further convolute the identification of valid accelerator configurations.
Constrains can arise from hardware limitations or if the evaluation of a configuration is impossible or too expensive~\cite{timeloop}.

To address the aforementioned challenges, we introduce a general accelerator architecture exploration framework, dubbed \apollo, that leverages the recent advances in black-box optimization to facilitate finding optimal design configurations under different design constraints.
We demonstrate how leveraging tailored optimization strategies for complex and high-dimensional space of architecture exploration yields large improvements (up to \speedupareafour) with a reasonably small number of evaluations ($\approx$ 0.0004\% of the search space).
Finally, we present the very first study on the impact of transfer learning between architecture exploration tasks with different design constraints in further reducing the number of hardware evaluations.
The following outlines the contributions of \apollo, making the first transferable architecture exploration infrastructure:
\begin{itempacked}
\item \niparagraph{End-to-end architecture exploration framework.}
We introduce and develop \apollo, an end-to-end and highly configurable framework for architecture exploration.
The proposed framework tunes accelerator configurations for a target set of workloads with a relatively small number of hardware evaluations.
As hardware simulations are generally time-consuming and expensive to obtain, reducing the number of these simulations not only shortens the design cycle for accelerators, but also provides an effective way to adapt the accelerator itself to various target workloads.
\item \niparagraph{Supporting various optimization strategies.}
\apollo introduces and employs a variety of optimization strategies to facilitate the analysis of optimization performance in the context of architecture exploration.
Our evaluations results show that evolutionary and population-based black-box optimization strategies yield the best accelerator configurations  (up to \speedupareafour speedup) compared to a baseline black-box optimization with only $\approx$ 2K number of hardware evaluations ($\approx$ \xx{0.0004\%} of search space). 
\item \niparagraph{Transfer learning for architecture exploration.}
Finally, we study and explore transfer learning between architecture exploration tasks with different design constraints showing its benefit in improving the optimization results and sample-efficiency.
Our results show that transfer learning not only improves the optimization outcome (up to \xx{25$\%$}) compared to independent exploration, but also reduces the number of hardware evaluations. 

\end{itempacked}
\section{Methodology}
\label{sec:methodology}

\niparagraph{Problem definition.}
The objective in \apollo (architecture exploration) is to discover a set of feasible accelerator parameters ($h$) for a set of workloads ($w$) such that a desired objective function ($f$), e.g. weighted average of runtime, is minimized under an optional set of user-defined constraints, such as area ($\alpha$) and/or runtime budget ($\tau$).

\begin{equation}
\begin{aligned}
\min_{h, w} \quad & f(h, w)\\
\textrm{s.t.} \quad & \mathrm{Area}(h) \leq \alpha \\
\quad \quad & \mathrm{Latency}(h, w) \leq \tau
\end{aligned}
\end{equation}
The manifold of architecture search generally contains infeasible points~\cite{prac_dse:mascots:2019}, for example due to impractical hardware implementation for a given set of parameters or impossible mapping of workloads to an accelerator.
As such, one of the main challenges for architecture exploration is to effectively sidestep these infeasible points.
We present and analyze the performance of optimization strategies to reduce the number of infeasible trials in Section~\ref{sec:eval}.

\niparagraph{Neural models.}
We evaluate \apollo on two variations of MobileNet~\cite{mnv2:arxiv:2018,edgetpu:arxiv:2020} models and five in-house neural networks with distinct accelerator resource requirements.
The neural model configurations, including their target domain, number of layers, and total filter sizes are detailed in Table~\ref{tab:models}.
In the multi-model study, the workload contains MobileNetV2~\cite{mnv2:arxiv:2018}, MobileNetEdge~\cite{edgetpu:arxiv:2020}, \mthree, \mfour, \mfive, \msix, and \mseven.
\begin{table}[ht]
  \begin{center}
    \caption{The detailed description of the neural models, their domains, number of layers, parameter size in megabytes, and number of MAC operations in million.}
    \label{tab:models}
    \begin{tabular}{l|l|l|l|l}
      \textbf{Name} & \textbf{Domain} & \textbf{\# of layers} & \textbf{Params (MB)} & \textbf{\# of MACs}\\
      \hline
      MobileNetV2~\cite{mnv2:arxiv:2018} & Image Classification & 76 & 3.33 & 301~M\\
      MobileNetEdge~\cite{efficientnet:2020} & Image Classification & 93 & 3.88 & 991~M\\
      \mthree & Object Detection & 93 & 2.19 & 464~M\\
      \mfour & Object Detection & 111 & 0.42 & 107~M\\
      \mfive & Object Detection & 60 & 6.29 & 1721~M\\
      \msix & Semantic Segmentation & 62 & 0.37 & 591~M\\
      \mseven & OCR & 56 & 0.30 & 5.19~M\\
    \end{tabular}
  \end{center}
\end{table}

\niparagraph{Accelerator search space.}
In this work, we use an in-house and highly parameterized edge accelerator.
The accelerator contains a 2D array of processing elements (PE) with multiple compute lanes and dedicated register files, each operating in single-instruction multiple-data (SIMD) style with multiply-accumulate (MAC) compute units.
There are distributed local and global buffers that are shared across the compute lanes and PEs, respectively. 
We designed a cycle-accurate simulator that faithfully models the main microarchitectural details and enables us to perform architecture exploration.
Table~\ref{tab:arch_params} outlines the microarchitectural parameters (e.g. compute, memory, or bandwidth) and their number of discrete values in the search space.
The total number of design points explored in \apollo is nearly $5\times10^8$.
\begin{table}[ht]
  \begin{center}
    \caption{The microarchitecture parameters, their type, and number of discrete values per parameter. The total number of design points per each study is 452,760,000.}
    \label{tab:arch_params}
    \begin{tabular}{l|c||l|c}
      \textbf{Accelerator Parameter} &\textbf{\# discrete values} & \textbf{Accelerator Parameter} & \textbf{\# discrete values}\\
      \hline
      \# of PEs-X & 10 & \# of PEs-Y & 10 \\
      Local Memory & 7 & \# of SIMD units & 7 \\
      Global Memory & 11 & \# of Compute lanes & 10 \\
      Instruction Memory & 4 & Parameter Memory & 5 \\
      Activation Memory & 7 & I/O Bandwidth & 6
    \end{tabular}
  \end{center}
\end{table}

\subsection{Optimization Strategies}
In \apollo, we study and analyze the performance of following optimization methods.

\niparagraph{Evolutionary.}
Performs evolutionary search using a population of $K$ individuals, where the genome of each individual corresponds to a sequence of discretized accelerator configurations. New individuals are generated by selecting for each individual two parents from the population using tournament selecting, recombining their genomes with some crossover rate $\gamma$, and mutating the recombined genome with some probability $\mu$. Following~\citet{real2019regularized}, individuals are discarded from the population after a fixed number of optimization rounds (‘death by old age’) to promote exploration. In our experiments, we use the default parameters $K=100$, $\gamma=0.1$, and $\mu=0.01$.

\niparagraph{Model-Based Optimization (MBO).}
Performs model-based optimization with automatic model selection following~\cite{angermueller2019model}. At each optimization round, a set of candidate regression models are fit on the data acquired so far and their hyper-parameter optimized by randomized search and five fold cross-validation. Models with a cross-validation score above a certain threshold are ensembled to define an acquisition function.
The acquisition is optimized by evolutionary search and the proposed accelerator configurations with the highest acquisition function values are used for the next objective function evaluation.

\niparagraph{Population-Based black-box optimization (P3BO).}
Uses an ensemble of optimization methods, including Evolutionary and MBO, which has been recently shown to increase sample-efficiency and robustness~\cite{p3bo:arxiv:2020}. Acquired data are exchanged between optimization methods in the ensemble, and optimizers are weighted by their past performance to generate new accelerator configurations. Adaptive-P3BO is an extension of P3BO which further optimizes the hyper-parameters of optimizers using evolutionary search, which we use in our experiments.

\niparagraph{Random.}
Samples accelerator configurations uniformly at random from the defined search space.

\niparagraph{Vizier.} 
An alternative approach to MBO based on Bayesian optimization with a Gaussian process regressor and the expected improvement acquisition function, which is optimized by gradient-free hill-climbing~\cite{vizier:sigkdd:2017}. Categorical variables are one-hot encoded.

We use the Google Vizier framework~\cite{vizier:sigkdd:2017} with the optimization strategies described above for performing our experiments.
We use the default hyper-parameter of all strategies~\cite{vizier:sigkdd:2017,p3bo:arxiv:2020}.
Each optimization strategy is allowed to propose 4096 trials per experiment. We repeat each experiment five times with different random seeds and set the reward of infeasible trials to zero. To parallelize hardware simulations, we use 256 CPU cores each handling one hardware simulation at a time. We further run each optimization experiment asynchronously with 16 workers that can evaluate up to 16 trials in parallel.

\section{Evaluation}
\label{sec:eval}

\begin{figure}[t]
    \centering
    \subfloat[\bench{MobileNetV2}]{
    \includegraphics[width=0.32\textwidth]{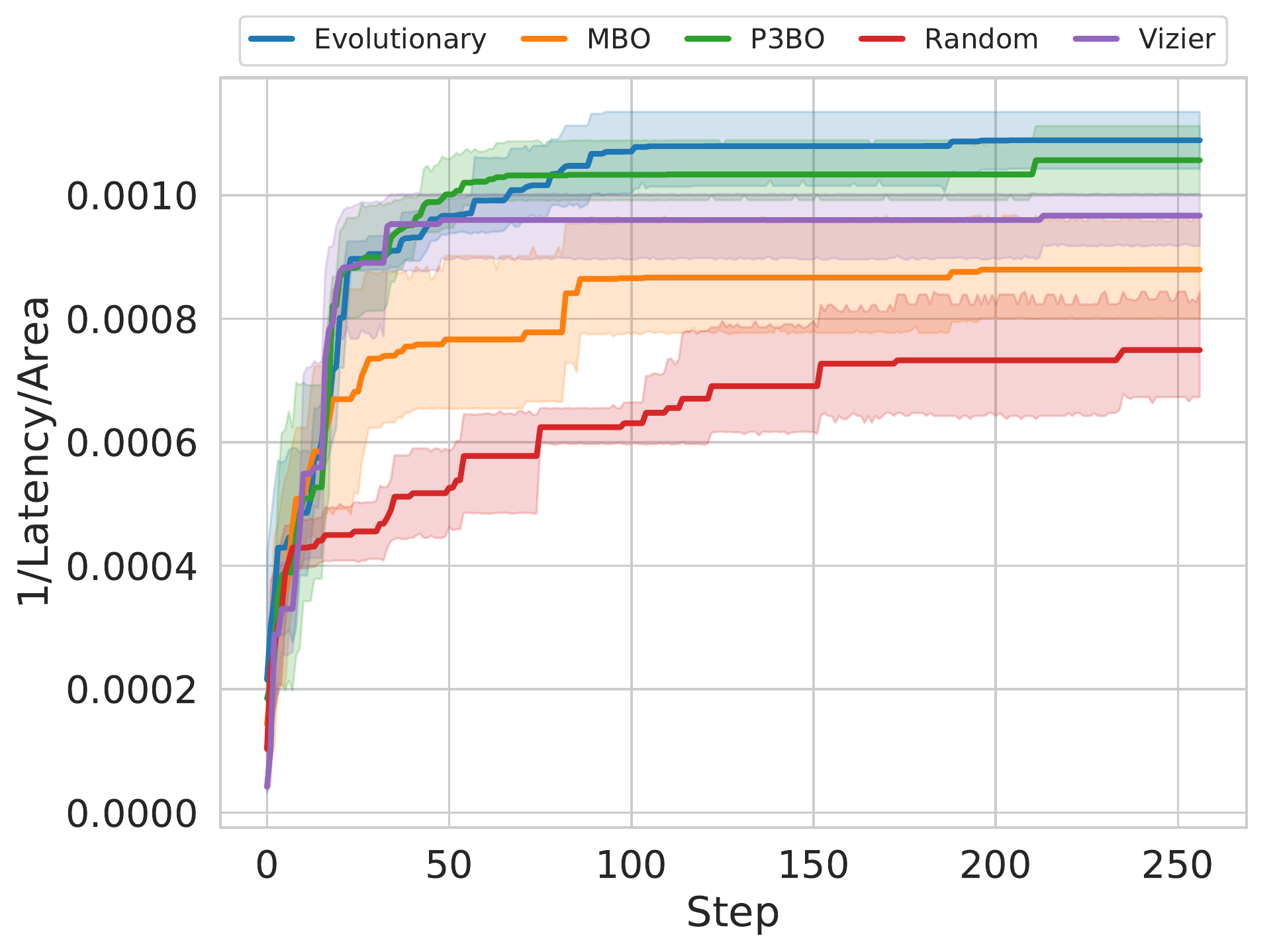}
    \label{fig:mnv2}}
    \subfloat[\bench{MobileNetEdge}]{
    \includegraphics[width=0.32\textwidth]{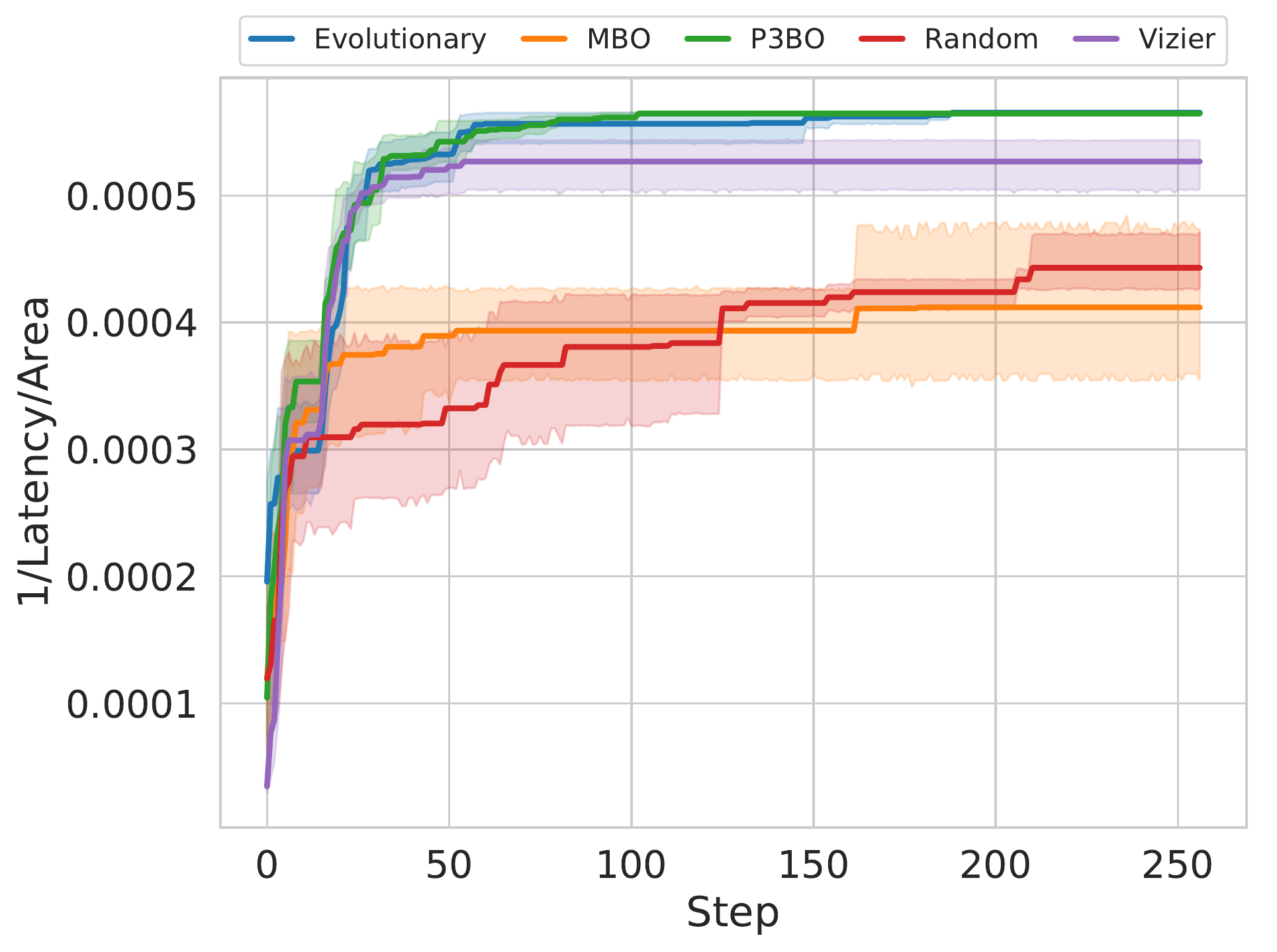}
    \label{fig:mnedge}}
     \subfloat[\bench{\mthree}]{
     \includegraphics[width=0.32\textwidth]{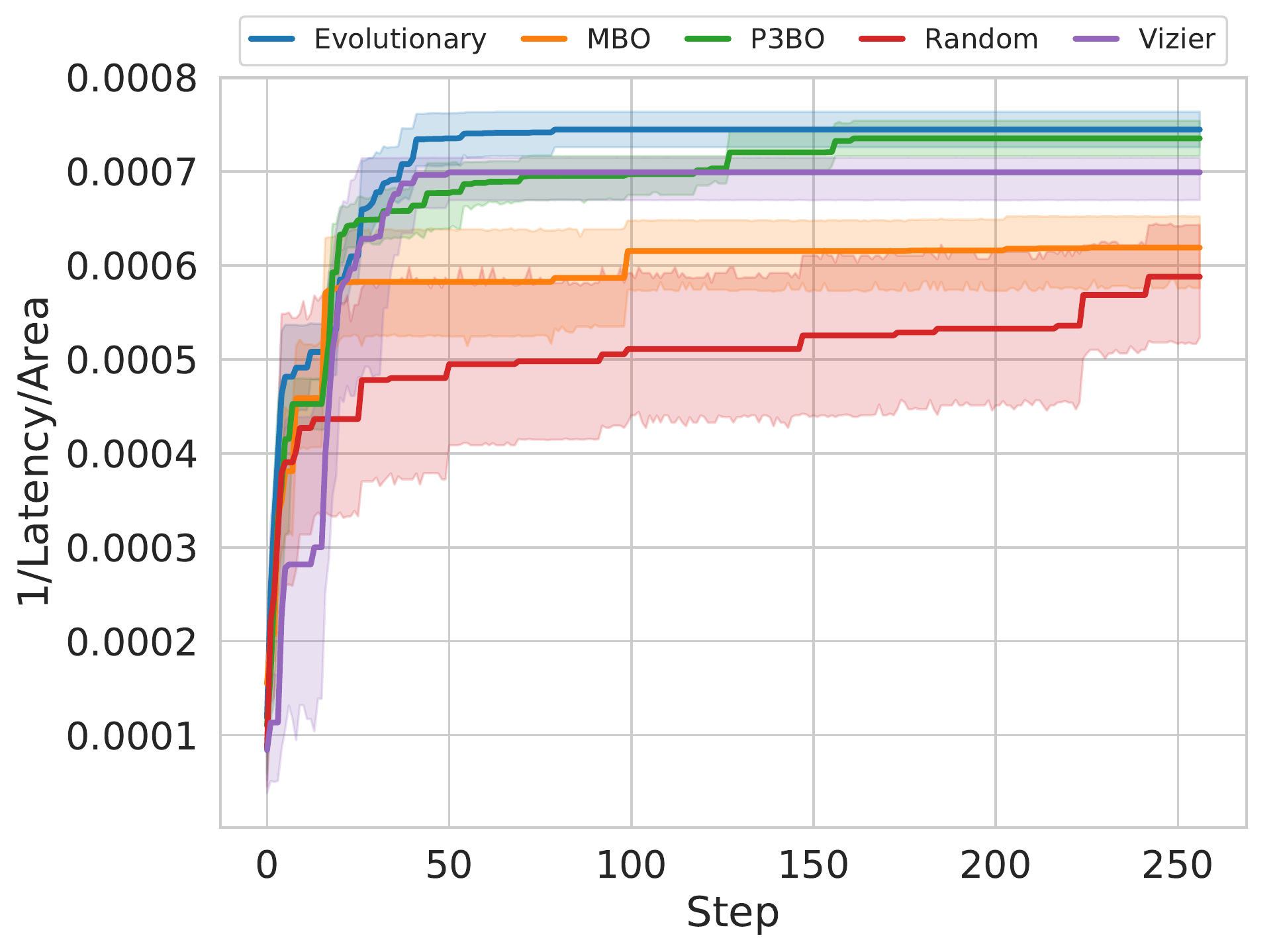}
    \label{fig:mthree}}
    \newline
    \subfloat[\bench{\mfour}]{
     \includegraphics[width=0.32\textwidth]{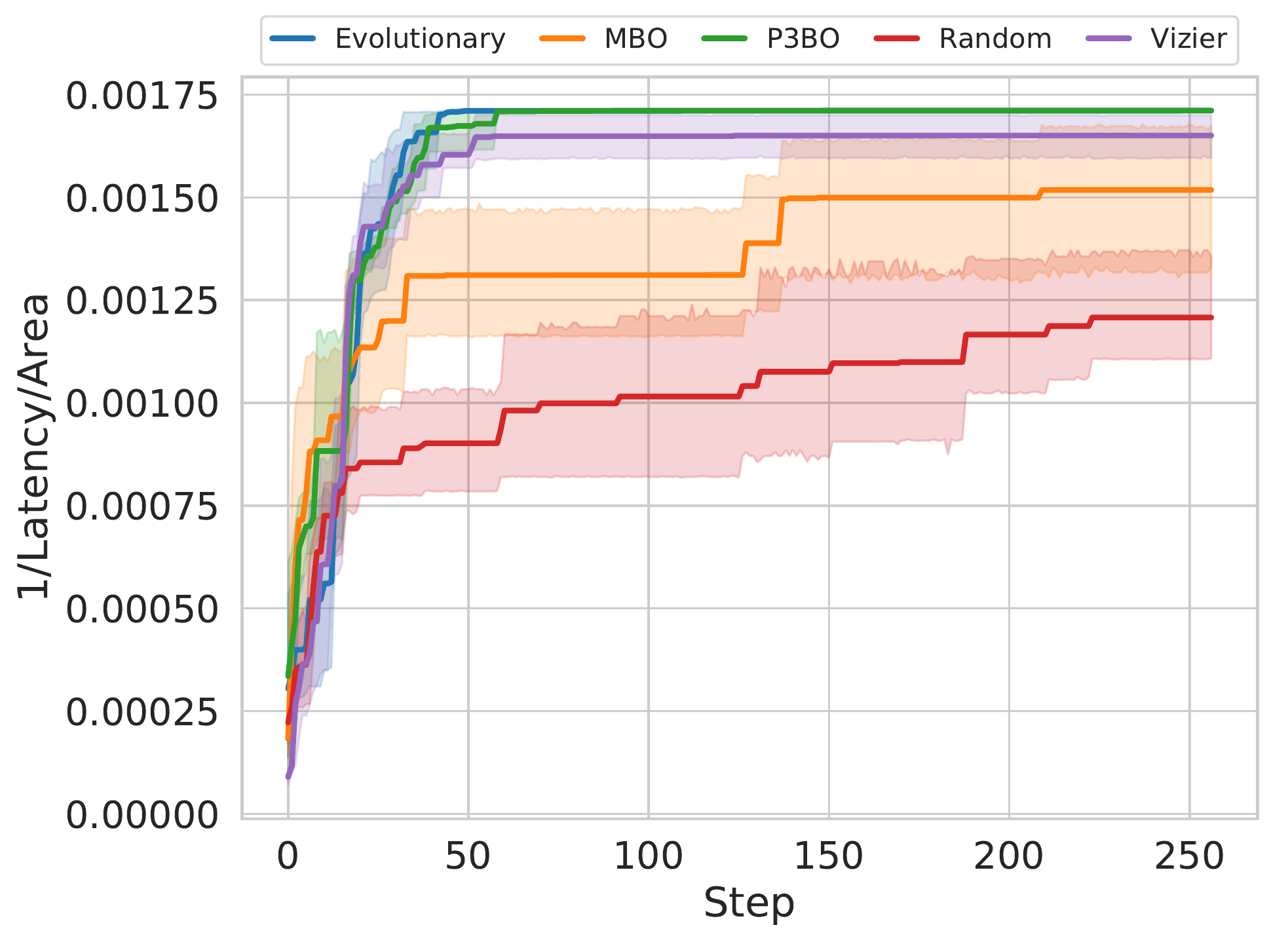}
    \label{fig:mfour}}
    \subfloat[\bench{\mfive}]{
     \includegraphics[width=0.32\textwidth]{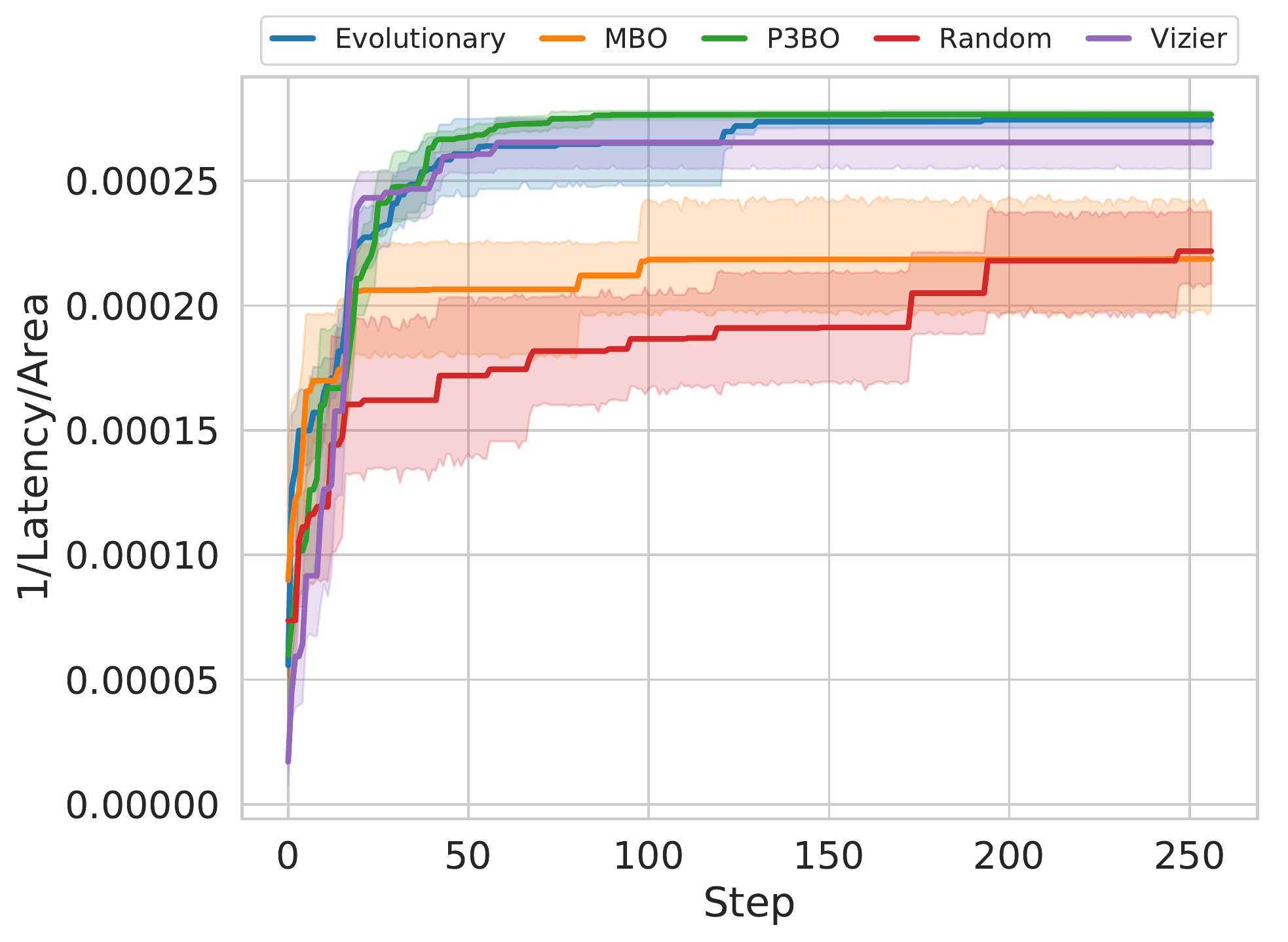}
    \label{fig:mfive}}
    \subfloat[\bench{\msix}]{
     \includegraphics[width=0.32\textwidth]{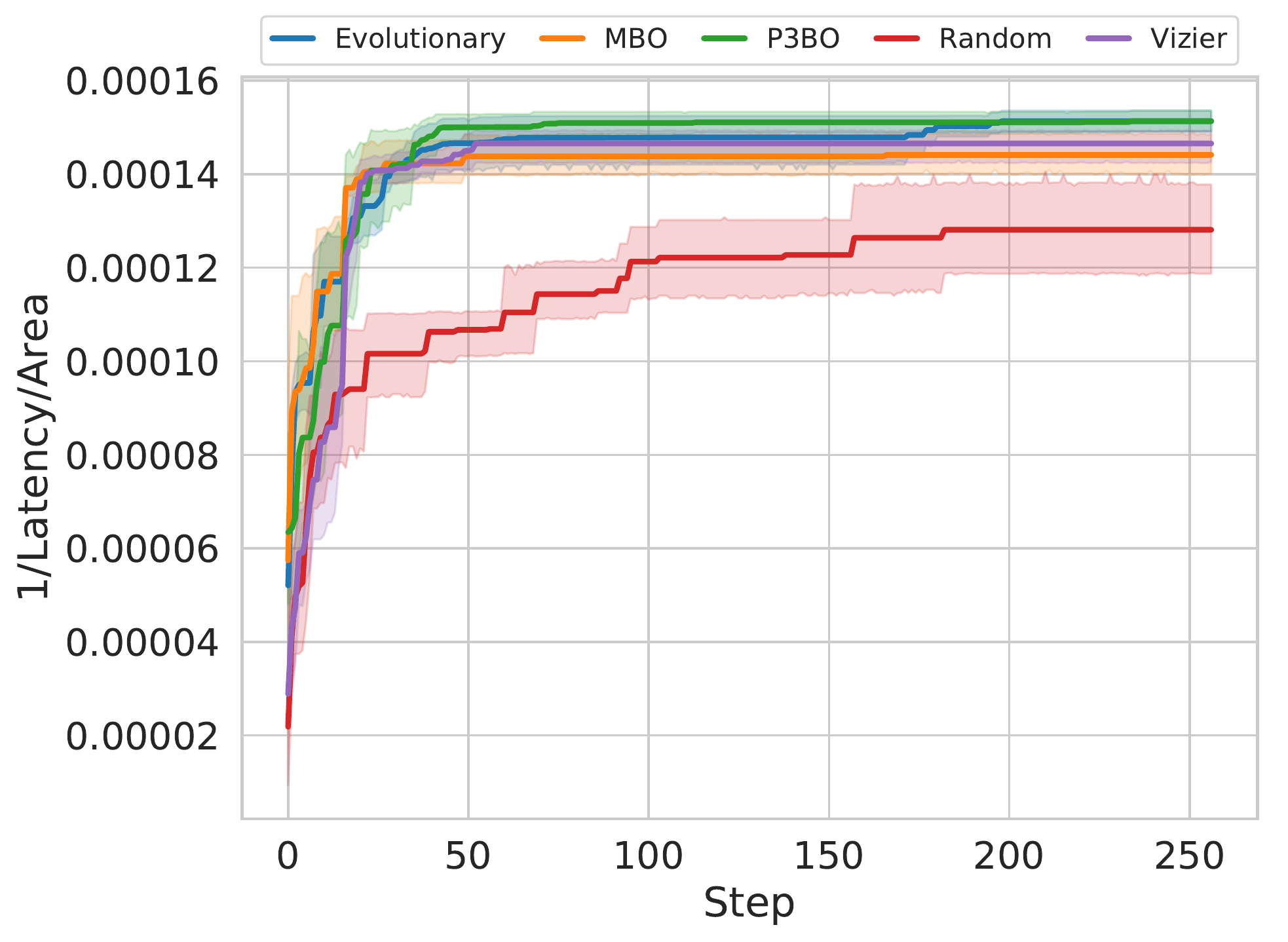}
    \label{fig:msix}}
    \caption{Performance of optimization strategies across various neural models in maximizing the throughput per area ($\frac{1}{latency}\times\frac{1}{area}$) ($\uparrow$ is better). The shaded area depicts the 95\% bootstrap confidence interval over five runs. Evolutionary and P3BO find high reward accelerator configurations faster than alternative optimization strategies.} 
    \label{fig:single_model}
    \vspace{-0.5cm}
\end{figure}

\niparagraph{Single model architecture search.}
For the first experiment, we define the optimization problem as maximizing throughput per area (e.g. $\frac{1}{latency}\times\frac{1}{area}$) for each neural model without defining any design constraints.
Figure~\ref{fig:single_model} depicts the cumulative reward across various number of trials.
Compared to \bench{Vizier}, \bench{Evolutionary} and \bench{P3BO} improve the throughput per area by \xx{4.3\%} (up to \xx{12.2\%} in \bench{MobileNetV2}), on average.
In addition, both \bench{Evolutionary} and \bench{P3BO} yield lower variance across multiple runs suggesting a more robust optimization method for architecture search.

\begin{figure}[t]
    \centering
    \subfloat[\bench{Area Budget = 6.8}]{
    \includegraphics[width=0.32\textwidth]{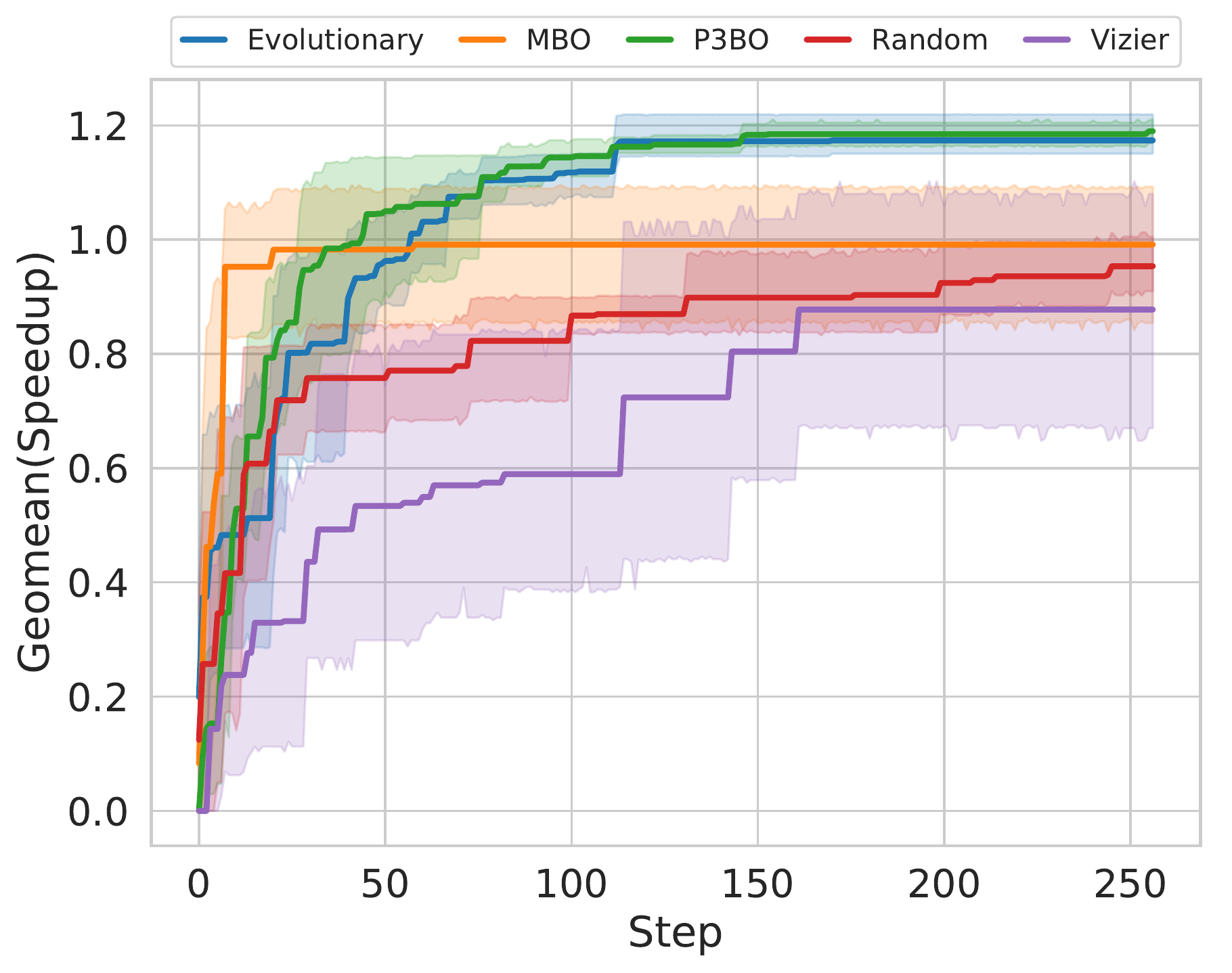}
    \label{fig:va_6.8}}
    \subfloat[\bench{Area Budget = 5.8}]{
    \includegraphics[width=0.32\textwidth]{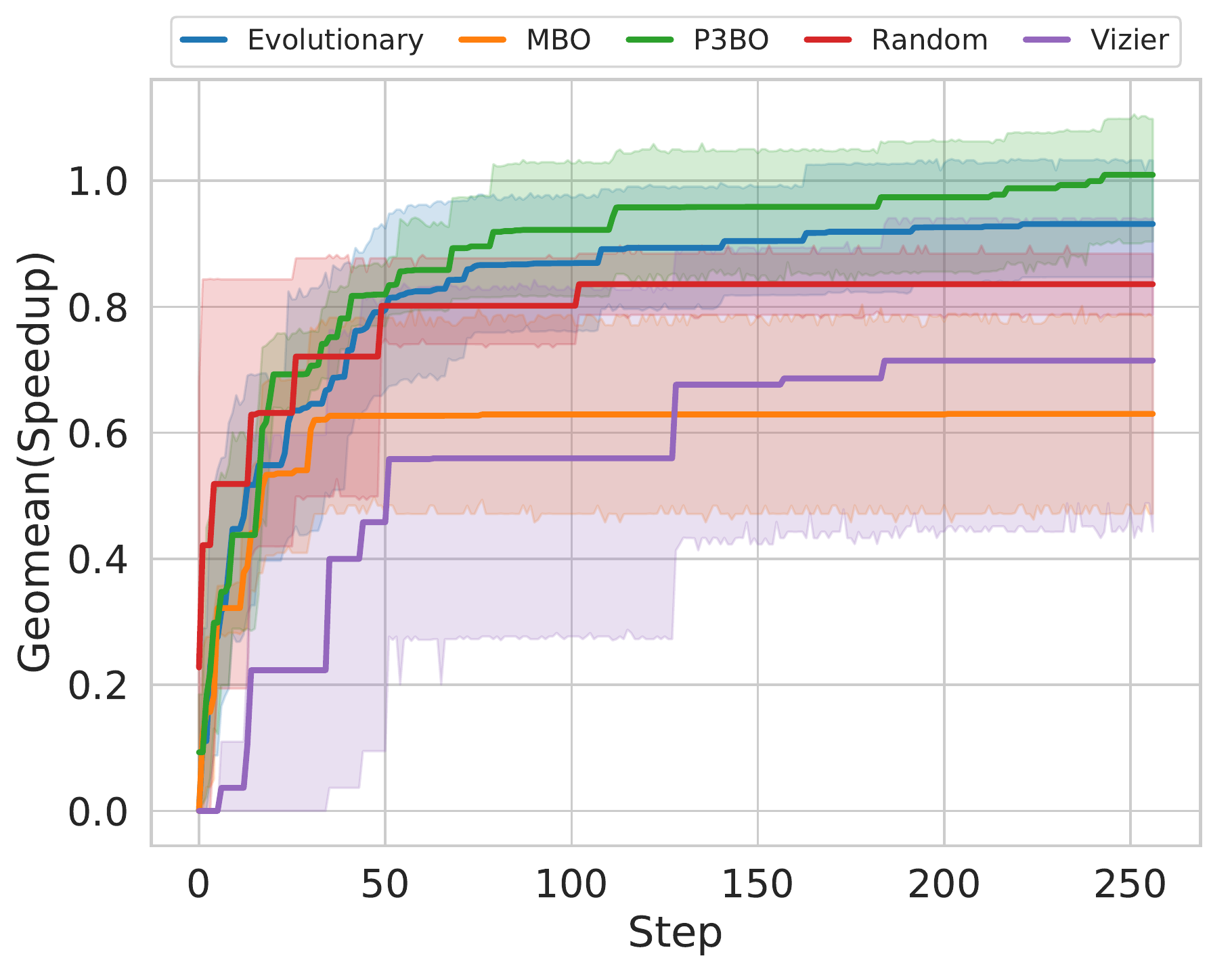}
    \label{fig:va_5.8}}
    \subfloat[\bench{Area Budget = 4.8}]{
    \includegraphics[width=0.32\textwidth]{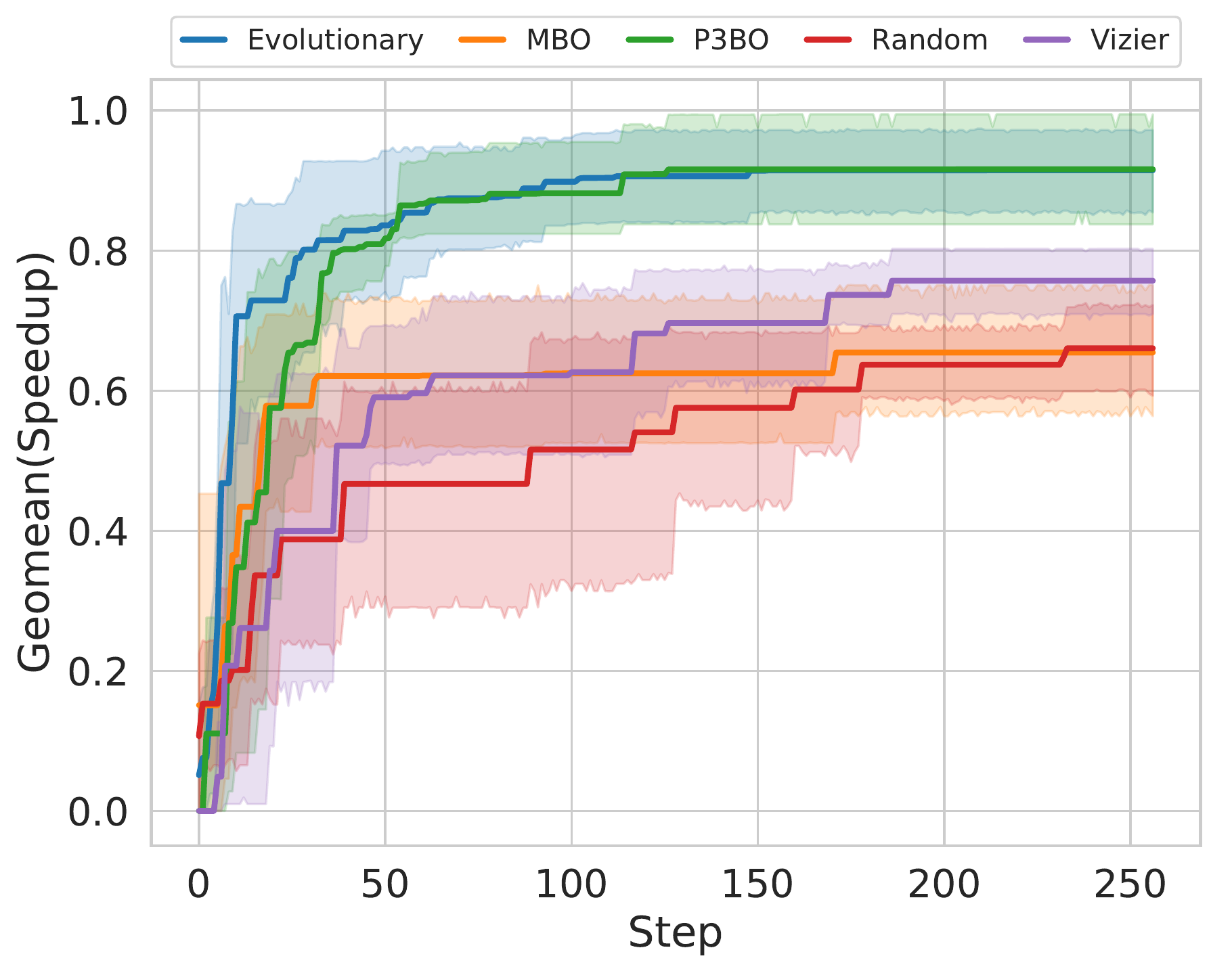}
    \label{fig:va_4.8}}
    \caption{Performance of optimization strategies in maximizing $\mathtt{geomean(speedup)}$ ($\uparrow$ is better) under alternative area budget constrains. The shaded area depicts the 95\% bootstrap confidence interval over five runs. The baseline latency numbers are from a productionized edge accelerator. As the area constraint becomes tighter (more infeasible points), the improvement by P3BO increases.}
    \label{fig:multi_model}
    \vspace{-0.5cm}
\end{figure}

\niparagraph{Multi-model architecture search.}
For multi-model architecture search, we define the optimization as maximizing \code{geomean(speedup)} across all the evaluated models (See Section~\ref{sec:methodology}) while imposing area budget constraints of 6.8 $\mathrm{mm}^{2}$, 5.8 $\mathrm{mm}^{2}$, and 4.8 $\mathrm{mm}^{2}$.
Note that, as the area budget becomes stricter, the number of infeasible trials increases.
The baseline runtime numbers are obtained from a productionized edge accelerator.
Figure~\ref{fig:multi_model} demonstrates the cumulative reward (e.g. \code{geomean(speedup)}) across various number of sampled trials.
Across the studied optimization strategies, \bench{P3BO} delivers the highest improvements across all the design constraints.
Compared to \bench{Vizier}, \bench{P3BO} improves the speedup by \speedupareasix, \speedupareafive, and \speedupareafour for area budget 6.8 $\mathrm{mm}^2$, 5.8 $\mathrm{mm}^2$, and 4.8 $\mathrm{mm}^2$, respectively.
These results demonstrate that as the design space becomes more constrained (e.g. more infeasible points), the improvement by \bench{P3BO} increases, showing its performance in navigating the search space better.

\niparagraph{Analysis of infeasible trials.}
To better understand the effectiveness of each optimization strategy in selecting feasible trials and unique trials, we define two metrics \textit{feasibility ratio} and \textit{uniqueness ratio}, respectively.
The feasibility (uniqueness) ratio defines the fraction of feasible (unique) trials over the total number of sampled trials.
Higher ratios generally indicate improved exploration of feasible regions.
Table~\ref{tab:feas_ratio} summarizes the feasibility and uniqueness ratio of each optimization strategy for area budget 6.8~$\mathrm{mm}^2$, averaged over multiple optimization runs.
\bench{MBO} yields the highest avg. feasibility ratio of \xx{$\approx$ 0.803} while \bench{Random} shows the lowest ratio of \xx{$\approx$ 0.009}.
While \bench{MBO} features a high feasibility ratio, it underperforms compared to other optimization strategies in finding accelerator configurations with high performance.
The key reason attributed to this behavior for \bench{MBO} is its low performance (\xx{0.236}) in identifying unique accelerator parameters compared to other optimization strategies.

\begin{table}[htbp]
  \begin{center}
    \caption{The average feasibility and uniqueness ratio across five runs for architecture search with an area budget of 6.8 $\mathrm{mm^2}$ (see Figure~\ref{fig:va_6.8}).}
    \label{tab:feas_ratio}
    \begin{tabular}{l||c|c|c|c|c}
      & \bench{Evolutionary} & \bench{MBO} & \bench{P3BO} & \bench{Random} & \bench{Vizier}\\
      \hline
      Avg. Feasibility Ratio ($\uparrow$ better)	& 0.362 & 0.803 & 0.347 & 0.009 & 0.012 \\
      Avg. Uniqueness Ratio ($\uparrow$ better) & 0.891 & 0.236 & 0.848 & 1.0 & 0.979
    \end{tabular}
  \end{center}
\end{table}

\begin{figure}[t]
    \centering
    \subfloat[tSNE of the 50 best configurations found by different methods (best viewed in color).]{
    \includegraphics[width=0.4\textwidth]{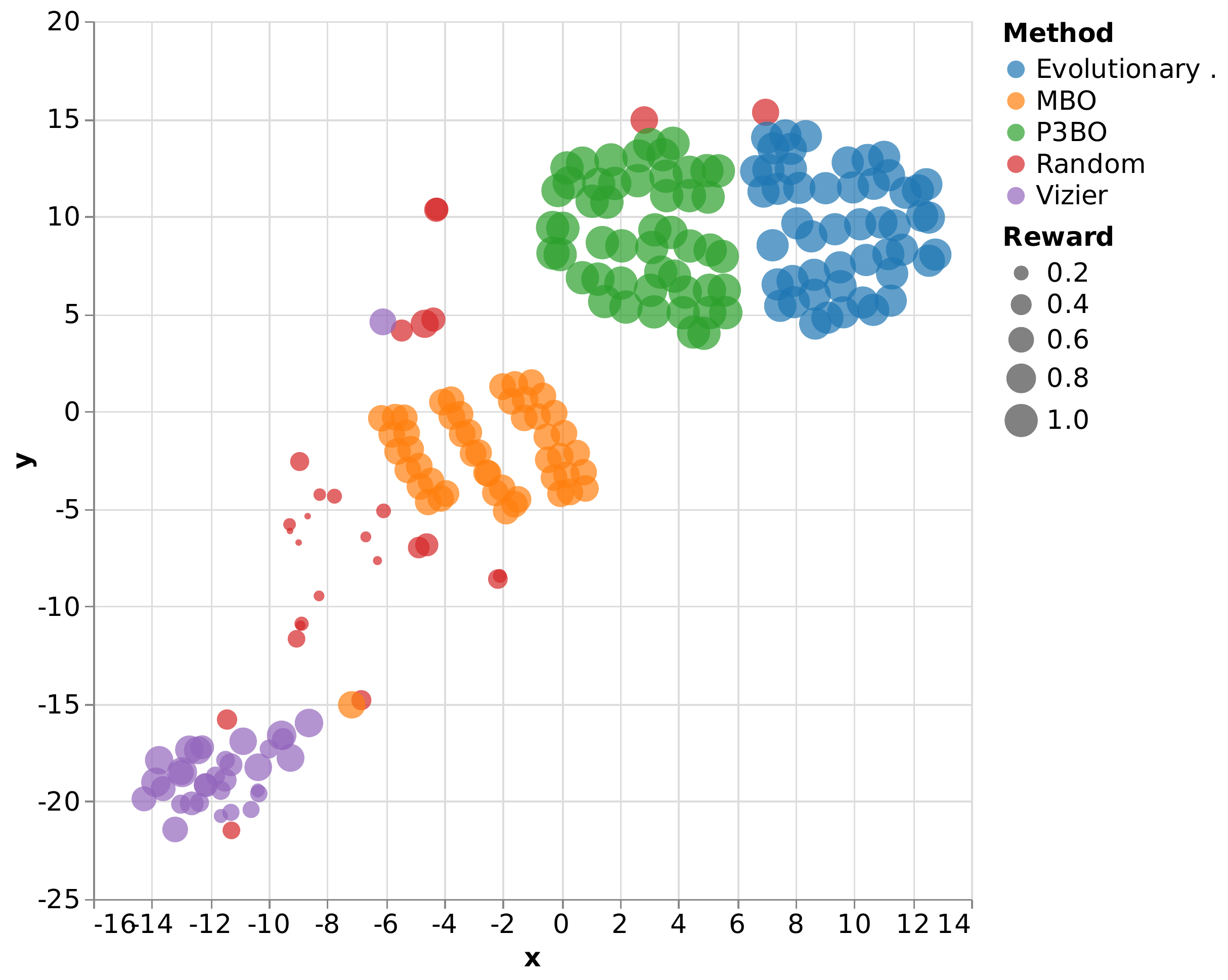}
    \label{fig:diversity_tsne}}
    \qquad
    \subfloat[Mean pairwise Euclidean distance of all configuration with a reward above the 75th percentile of the maximum reward. Error bars show the variance across 5 optimization runs.]{
    \includegraphics[width=0.42\textwidth]{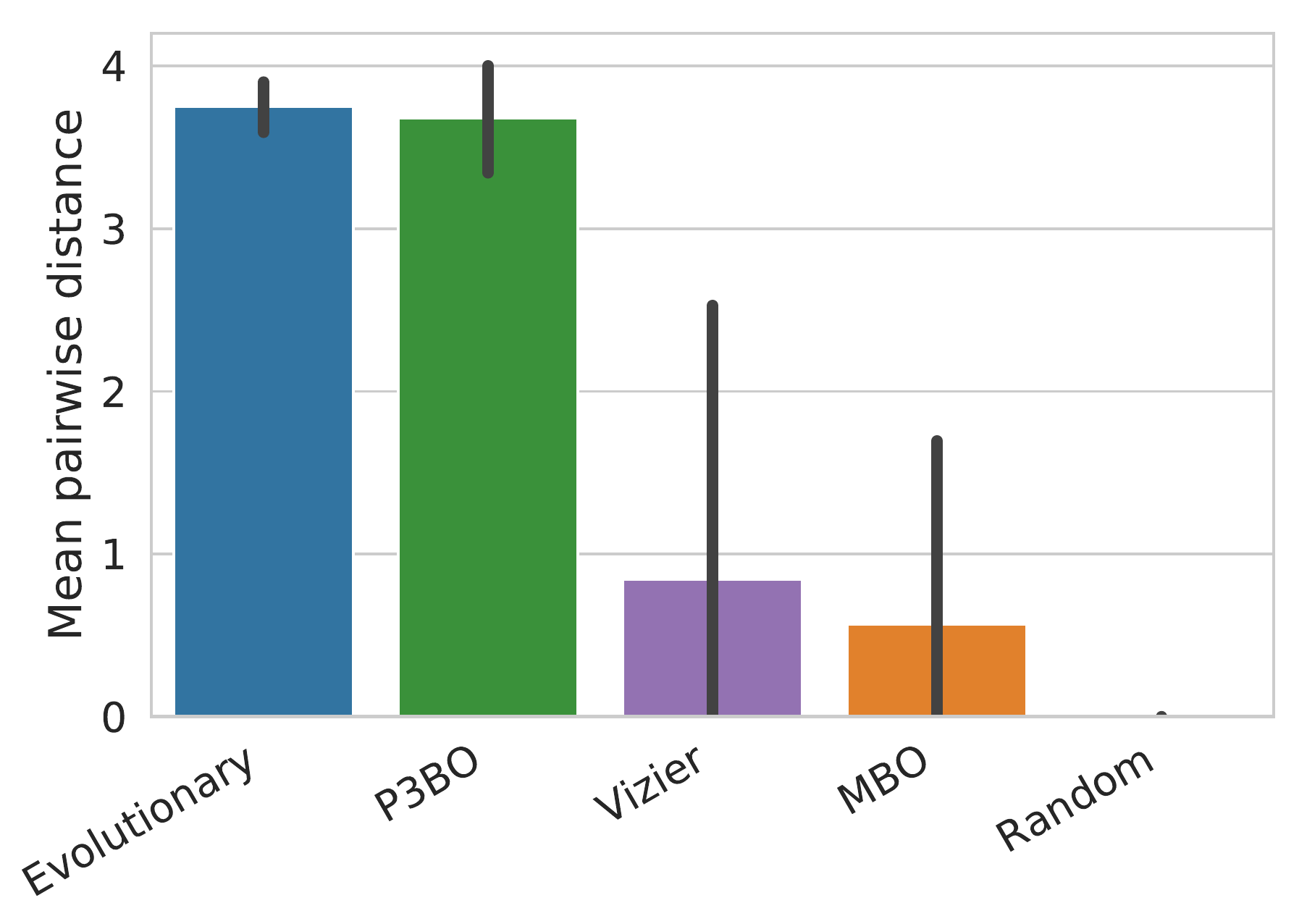}
    \label{fig:diversity_mean_pairwise}}
    \caption{Diversity quantification of architecture configurations found by different methods for an area budget of 4.8 $\mathrm{mm}^2$.}
    \label{fig:diversity}
\end{figure}

\begin{figure}[t]
    \centering
    \includegraphics[width=0.45\textwidth]{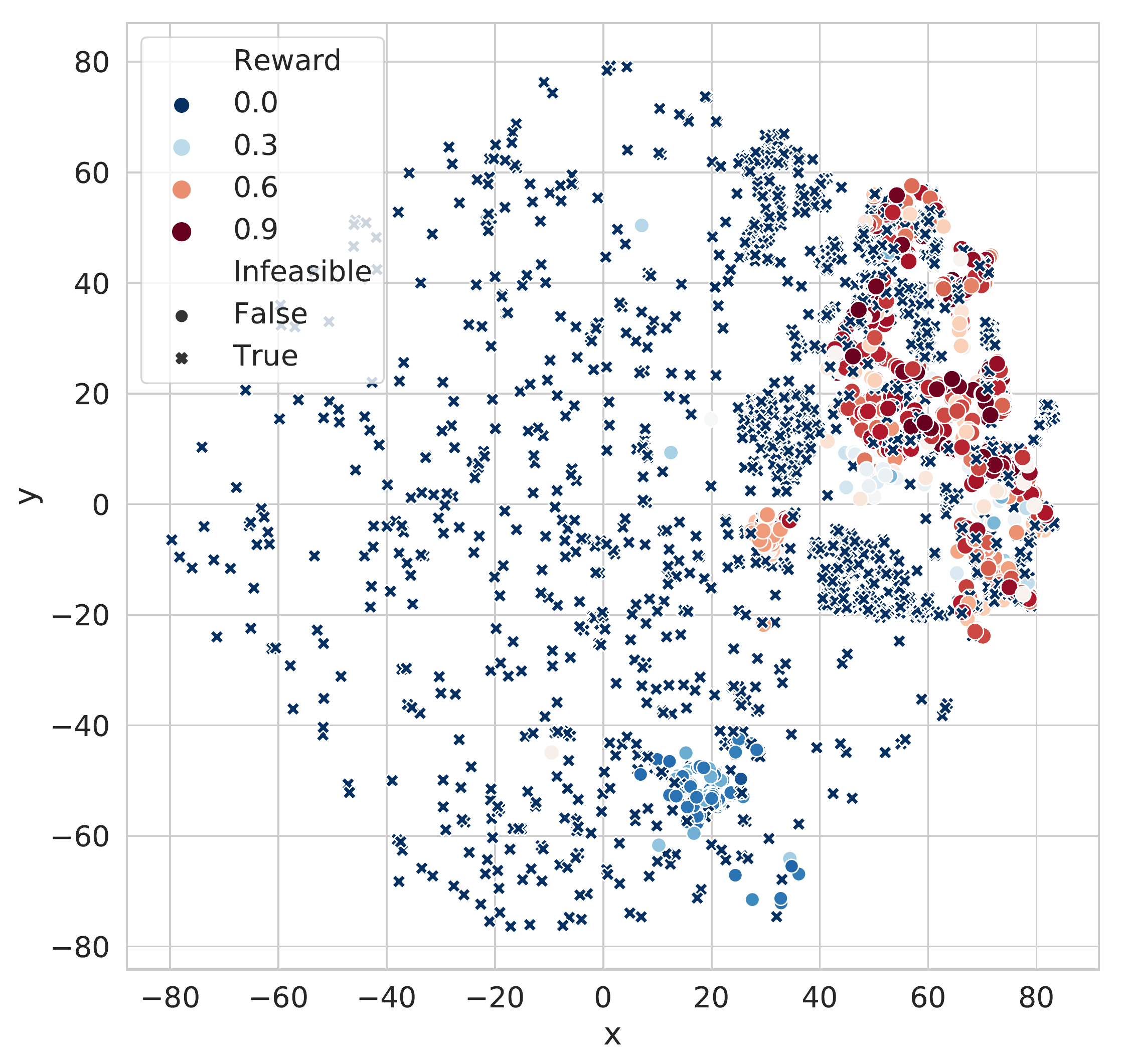}
    \caption{tSNE of all trials (including infeasible ones) proposed by the Evolutionary algorithm for an area budget of 4.8 $\mathrm{mm}^2$. Shows the large fraction of infeasible trials (crosses) vs feasible (circles) trials (best viewed in color).}
    \label{fig:diversity_infeasible_tsne}
\end{figure}

\niparagraph{Diversity of architecture configurations.}
A desired property of optimizers is to not only find a single but a diverse set of architecture configurations with a high reward that can be tested downstream. We quantified the ability of optimizers to find diverse configurations qualitatively by visualizing the 50 best unique trials found by each method using tSNE. Figure~\ref{fig:diversity_tsne} shows that \bench{Evolutionary} and \bench{P3BO} find both higher-reward and more diverse configurations compared to alternative methods with the exception of \bench{Random}. This finding is supported quantitatively by Figure~\ref{fig:diversity_mean_pairwise}, which shows the mean pairwise Euclidean distance of configurations with a reward above the 75th percentile of the maximum reward. The mean pairwise distance of \bench{Random} is zero since it did not find any configurations with a reward above the 75th percentile.
To further visualize the search space in architecture exploration, Figure~\ref{fig:diversity_infeasible_tsne} shows the tSNE visualization of all trials proposed by the \bench{Evolutionary} method for an area budget of 4.8~$\mathrm{mm}^2$.
This figure shows the large number of infeasible trials in the space and the proximity of low- and high-performing trials, which renders identifying high-performing trials challenging.

\begin{figure}[t]\label{fig:tpo}
    \centering
    \includegraphics[width=0.4\textwidth]{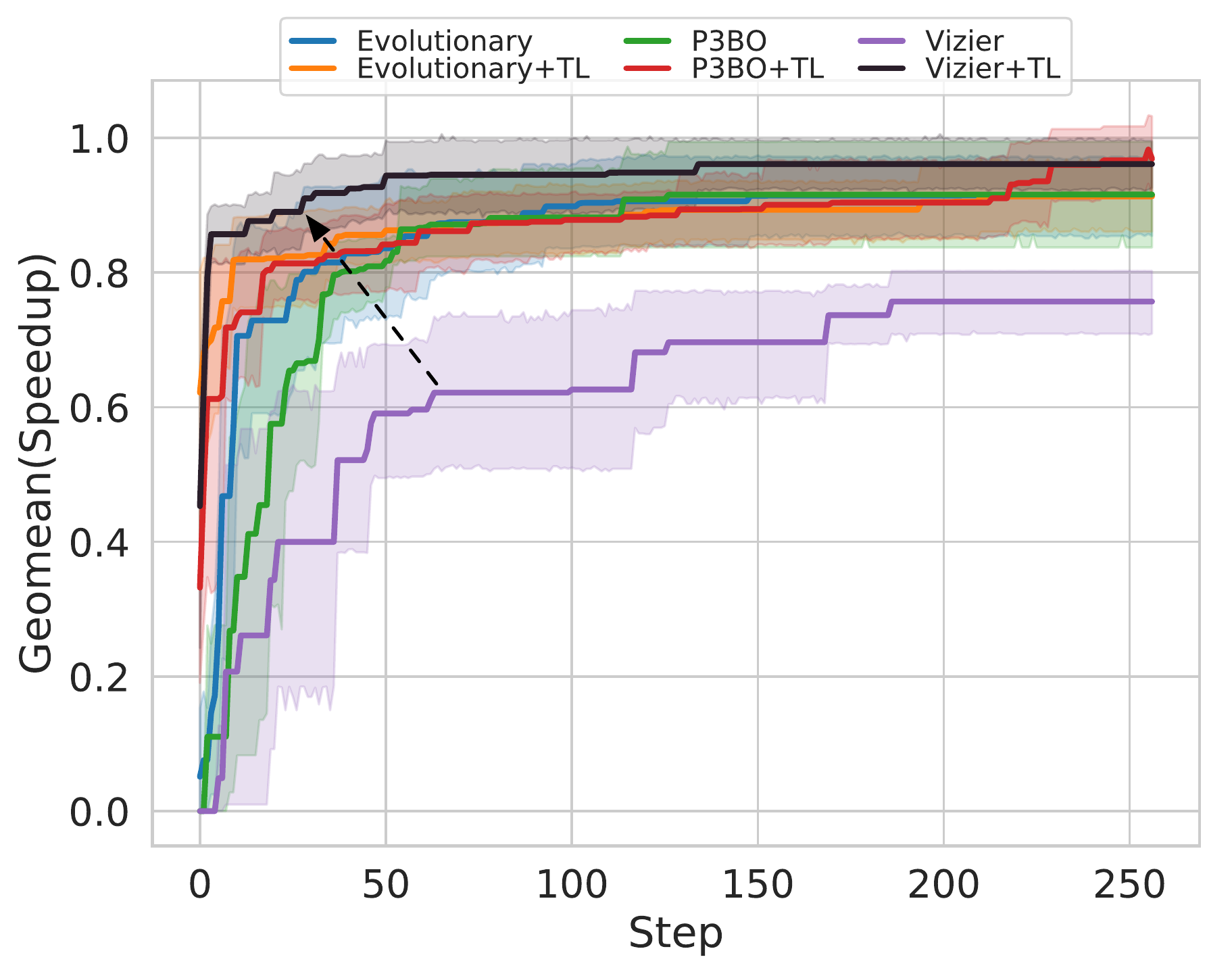}
    \caption{Comparing optimization strategies in maximizing the $\mathtt{geomean(speedup)}$ ($\uparrow$ is better) without transfer learning (top row in the legend) and with transfer learning (bottom row in the legend) for area budget 4.8~$\mathrm{mm}^2$. Transfer learning enables finding higher performance accelerator configurations in fewer steps.}
    \label{fig:tl}
\end{figure}

\niparagraph{Transfer learning between optimizations with different constraints.}
We analyze the effect of transfer learning between architecture search tasks with different area budgets.
To create the source tasks, we select 100 unique trials from optimization studies with area budget constraint of 6.8~$\mathrm{mm}^2$ (See Fig.~\ref{fig:va_6.8}) under two criteria.
First, the area consumption of the selected trials must satisfy the area budget ( 4.8~$\mathrm{mm}^2$) of the target task.
Second, the objective function value (reward) of the selected trials must be below a predefined threshold.
In our experiments, we create two source tasks with an objective value of 0.8 and 0.4, respectively, which we chose to better understand the impact of low- and high-value rewards.
We use the selected trials to seed the optimization of the target task, which has an area budge of 4.8~$\mathrm{mm}^2$).
Figure~\ref{fig:tl} shows the results.
All the optimization strategies find high reward trials in fewer steps with transfer learning than without. The improvement is most pronounced for \bench{Vizier}, which finds trials with a reward of $\approx$ 1.0 with transfer learning compared to only $\approx$ 0.8 without transfer learning. This suggest that \bench{Vizier} uses the selected trials from the source task more efficiently than \bench{Evolutionary} and \bench{P3BO} for optimizing the target task.  

In our implementation, \bench{Evolutionary} and \bench{P3BO} simply use the 100 unique and feasible trails from the source task to initialize the population of evolutionary search.
Instead, \bench{Vizier} uses a more advanced transfer learning approach based on a stack of Gaussian process regressors (see Section 3.3 of \citet{vizier:sigkdd:2017}), which may account for the performance improvement.
We leave extending \bench{Evolutionary} and \bench{P3BO} by more advanced transfer learning approaches as future work.

\niparagraph{Comparison to exhaustive exploration.}
To understand the optimal design point, we perform a semi-exhaustive search within the search space. Since the search space has almost $5\times10^8$ design points, it is merely not practical to perform a fully-exhaustive search.
As such, we manually prune the search space using domain knowledge where the design points are within a typical edge accelerator configuration (e.g. total memory size within 4--16 MB, total number of PEs within 2--16, etc.).
Additionally, we perform a cheaper area estimation to reject design points before performing expensive cycle-level simulations.
Using this pruning approach, we reduced the size of search space to around 3K samples.
We observe that \bench{P3BO} can reach the best configurations found by the semi-exhaustive search by performing far fewer evaluations (\xx{1.36$\times$} less).
Another interesting observation is that for the multi-model experiment targeting 6.8~$\mathrm{mm}^2$, \bench{P3BO} actually finds a design slightly better than semi-exhaustive with 3K-sample search space. We observe that the design uses a very small memory size (3MB) in favor of more compute units. This leverages the compute-intensive nature of vision workloads, which was not included in the original semi-exhaustive search space.
This demonstrates the need of manual search space engineering for semi-exhaustive approaches, whereas learning-based optimization methods leverage large search spaces reducing the manual effort.

\section{Related Work}
\label{sec:related}
While inspired by related work, \apollo is fundamentally different from classic methodologies in design space exploration: (1)
we develop a platform to compare the effectiveness of a wide range of optimization algorithms; and (2)
we are the first work, to the best of our knowledge, that leverages transfer learning between architecture exploration tasks with different design constraints showing how transfer learning slashes the time for discovering new accelerator configurations.
Related work to \apollo embodies three broad research categories of black-box optimization, architecture exploration, and transfer learning. Below, we overview the most relevant work in these categories.

\niparagraph{Black-box optimization.}
Black-box optimization has been broadly applied across different domains and appeared under various optimization categories, including Bayesian ~\cite{pbo:arxiv:2020,p3bo:arxiv:2020,con_bo:ba:2019,human_bo:ieee:2015,population:is:2014,practical_bo:nips:2012,tpe:nips:2011,bo_tutorial:arxiv:2010}, evolutionary~\cite{survey_gen:jrs:2019,ssl_pso:gec:2018,eff_global:jgo:1998}, derivative-free~\cite{derivative_methods:arxiv:2019,derivative_review:jgo:2013,intro_derivative:siam:2009}, and bandit~\cite{bouneffouf:arxiv:2019,hyperband:jmlr:2017,gauss_bandit:arxiv:2009,surface:jrss:1995}.
\apollo benefits from advances in black-box optimization and establishes a basis for leveraging this broad range of optimization algorithms in the context of accelerator design. 
In this work, we extensively studied the effectiveness of some of these black-box optimization algorithms, namely random search~\cite{vizier:sigkdd:2017}, Bayesian optimization~\cite{vizier:sigkdd:2017}, evolutionary algorithms~\cite{p3bo:arxiv:2020}, and ensemble methods~\cite{p3bo:arxiv:2020} in discovering optimal accelerator configurations under different design objectives and constraints.

\niparagraph{Design space exploration.}
Design space exploration in computer systems has been always an active research and has become even more crucial due to the surge of specialized hardware~\cite{bo:frontiers:2020,flexibo:arxiv:2020,cnn_gen:cyber:2020,prac_dse:mascots:2019,accel_gen:dac:2018,spatial:pldi:2018,automomml:hpc:2016,opentuner:pact:2014}.
Hierarchical-PABO~\cite{bo:frontiers:2020} and FlexiBO~\cite{flexibo:arxiv:2020} use multi-objective Bayesian optimization for neural network accelerator design.
In order to reduce the use of computational resources, Sun et al.~\cite{cnn_gen:cyber:2020} apply genetic algorithm to design CNN models without modifying the underlying architecture.
HyperMapper~\cite{prac_dse:mascots:2019} uses a random forest in the automatic tuning of hardware accelerator parameters in a multi-objective setting.
HyperMapper optionally uses continuous distributions to model the search space variables as a means to inject prior knowledge into the search space.

\niparagraph{Transfer learning.}
Transfer learning exploits the acquired knowledge in some tasks to facilitate solving similar unexplored problems more efficiently, e.g. consuming a fewer number of data samples and/or outperforming previous solutions.
Transfer learning has been explored extensively and applied to various domains~\cite{min_multiproblem:ec:2020,metalearn:arxiv:2020,transfer:expert:2019,aml:springer:2019,transfer_mobj:ec:2017,trans_evol:cvprw:2017,regret_transfer:aistats:2017,transfer_ci_survery:know:2015,genetic_transfer:expert:2010, swersky2013multi}.
Due to the expensive-to-evaluate nature of hardware evaluations, transfer learning seems to be a practical mechanism for architecture exploration.
However, using transfer learning for architecture exploration and accelerator design is rather less explored territory.
\apollo is one of the first methods to bridge this gap between transfer learning and architecture exploration.

\section{Conclusion}
\label{sec:conclusion}
In this paper, we propose \apollo, a framework for sample-efficient architecture exploration for large scale design spaces.
The benefits of \apollo are most noticeable when architecture configurations are costly to evaluate, which is a common trait in various architecture optimization problems. 
Our framework also facilitates the design of new accelerators with different design constraints by leveraging transfer learning.
Our results indicate that transfer learning is effective in improving the target architecture exploration, especially when the optimization constraints have tighter bounds.  
Finally, we show that the evolutionary algorithms used in this work yield more diverse accelerator designs compared to other studied optimization algorithms, which can potentially discover overlooked architectures.
Architecture exploration is just one use case in the accelerator design process that is bolstered by \apollo.
The evolution of accelerator architectures mandates broadening the scope of optimizations to the entire computing stack, including scheduling and mapping, that potentially yields higher benefits at the cost of handling more complex optimization problems.
We argue that such co-evolution between the cascaded layers of the computing stack is inevitable in designing efficient accelerators honed for a diverse category of applications. This is an exciting path forward for future research directions.

\section{Acknowledgement}
\label{sec:ack}
We would like to thank Sagi Perel, Suyog Gupta, Cliff Young, David Dohan, D. Scully, and the Vizier team for their help and support.

\bibliographystyle{style/neurips_2020}
\bibliography{paper}

\end{document}